\documentclass[11pt,a4paper]{article}
\usepackage[hyphens]{url}
\usepackage{hyperref}
\usepackage[hyphenbreaks]{breakurl}
\usepackage[hyperref]{eacl2021}
\usepackage{times}
\usepackage{latexsym}
\usepackage{graphicx}

\usepackage{microtype}

\aclfinalcopy % Uncomment this line for the final submission
\title{Comparing Approaches to Dravidian Language Identification}

\author{Tommi Jauhiainen\textsuperscript{1,3}, Tharindu Ranasinghe\textsuperscript{2}, Marcos Zampieri\textsuperscript{3} \\
  \textsuperscript{1}University of Helsinki, Finland \\
  \textsuperscript{2}University of Wolverhampton \\
  \textsuperscript{3}Rochester Institute of Technology, USA \\
  \texttt{tommi.jauhiainen@helsinki.fi} \\}

\date{}

\begin{document}
\maketitle
\begin{abstract}

This paper describes the submissions by team HWR to the Dravidian Language Identification (DLI) shared task organized at VarDial 2021 workshop. The DLI training set includes 16,674 YouTube comments written in Roman script containing code-mixed text with English and one of the three South Dravidian languages: Kannada, Malayalam, and Tamil. We submitted results generated using two models, a Naive Bayes classifier with adaptive language models, which has shown to obtain competitive performance in many language and dialect identification tasks, and a transformer-based model which is widely regarded as the state-of-the-art in a number of NLP tasks. Our first submission was sent in the closed submission track using only the training set provided by the shared task organisers, whereas the second submission is considered to be open as it used a pretrained model trained with external data. Our team attained shared second position in the shared task with the submission based on Naive Bayes. Our results reinforce the idea that deep learning methods are not as competitive in language identification related tasks as they are in many other text classification tasks.

\end{abstract}

\section{Introduction}

Discriminating between similar languages (e.g. Bulgarian and Macedonian or Croatian and Serbian), language varieties (e.g. Brazilian and European Portuguese), and dialects is one of the main challenges in automatic language identification (LI) in texts. This issue has been addressed in a couple recent surveys \cite{jauhiainen2019survey,zampieri2020survey} and evaluation papers \cite{dslrec:2016}. Furthermore, it has been the topic of a number of competitions such as TweetLID \cite{zubiaga2016tweetlid} and many shared tasks organized in past VarDial workshops \cite{zampieri-EtAl:2018:VarDial,vardial2019report,vardial2020report}.

In this paper, we revisit the challenge of discriminating between similar languages on a challenging code-mixed data set containing three South Dravidian Languages: Kannada, Malayalam, and Tamil. In addition, there were code-mixed comments on other languages as well. The data set contains 22,164 YouTube comments divided into 16,674 instances for training and 4,590 instances for testing, each containing a mix of languages. The data set was released by the organizers of the Dravidian Language Identification (DLI) shared task part of the VarDial Evaluation Campaign 2020 \cite{vardial2021report}.\footnote{\url{https://sites.google.com/view/vardial2021/evaluation-campaign}} The goal of the task is to train computational models able to identify the language of each comment in a test set presented in Section \ref{sec:Data}. 

The DLI debuted as the first competition on similar language identification using code-mixed data associated with the VarDial workshops. Very similar multilingual language identification shared tasks were organized as part of Forum for Information Retrieval Evaluation (FIRE) meetings in 2013 and 2014 \citep{roy1,choudhury1}.\footnote{\url{http://fire.irsi.res.in/fire/2021/home}} A similar competition to the DLI are also the shared tasks on Language Identification in Code-Switched Data \cite{solorio2014overview,molina1}. These shared tasks addressed intra-sentential language identification with word level labeling whereas the DLI shared task is language identification shared task with predictions at the document level. We take this opportunity to evaluate the performance of two models for this task, a Naive Bayes classifier using adaptive language models and a transformers-based system described in detail in Section \ref{sec:Methods}. 

\section{Related Work}
\label{sec:RW}

The task of automatic language identification of texts has been under continuous research since 1960s as is witnessed, for example, by the works of \citet{mustonen1}, \citet{house1}, \citet{henrich1}, \citet{grefenstette1}, and \citet{martins1}. The problem of discriminating between similar languages, language varieties, and dialects is a particularly challenging one and it has also been addressed by a number of studies such as \citet{tiedemann2012efficient,tan:2014:BUCC,malmasi-zampieri:2017:VarDial1,malmasi-zampieri:2017:VarDial2}, and the aforementioned shared tasks at the VarDial workshop. In addition to addressing the issue of discriminating between similar languages, \citet{jauhiainen2019survey} provide an extensive overview of the history and methods used in language identification of texts in general.

\subsection{Language Identification of South Dravidian Languages}

Even though the DLI 2021 shared task was the first time a shared task solely focused on discriminating between Dravidian languages, Dravidian languages have been part of language repertoire of LI research before. Most of the research so far has focused on texts written with non-Roman script. 
%The native scripts of the Dravidian languages have evolved from the Brahmi script \citep{singh2010modeling} even though Sanskrit is not regarded as their ``mother'' language \citep{sreejith1}. 
Notable exception to the trend is the first subtask of the Transliterated Search shared which was organized as part of FIRE 2014 \citep{choudhury1}, where the goal was to label individual transliterated words in code-mixed search queries. The training set of the subtask included code-mixed sentences in English together with one of the Indian languages: Bangla, Gujarati, Hindi, or Malayalam. In addition, the test set included code-mixed sentences with Kannada and Tamil as unseen languages. \citet{choudhury1} list three teams which submitted results for the Dravidian languages, but to the best of our knowledge there are no system description papers available for these submissions.

Using Multivariate Analysis (MVA) and Principal Component Analysis (PCA), \citet{vinosh1} attained a 100\% performance in language identification between six Tamil dialects (Iscii, Shree-Tam, Tab, Tam, Tscii, and Vikatan). The exact test sizes they used are not known, but they considered 500 bytes ``too small'', which indicates that the texts they processed were generally much longer than those part of the DLI shared task test set.

\citet{murthy1} focused on smaller text samples in pair wise language identification of several Indian languages, including those of Tamil, Malayalam, and Kannada with Multiple Linear Regression (MLR). They used the number of aksharas (a sort of syllables generally used in Indian scripts) as the measure of the length of text and obtained F-scores of over 0.99 with texts of only 10 aksharas in length when discriminating between two of the South Dravidian languages. In their research, they show that features based on aksharas (\textit{e.g.} akshara \textit{n}-grams) work better than features based on bytes.

\citet{goswami2020unsupervised} experiment with supervised and unsupervised methods in dialect identification using, among others, a Dravidian data set containing Tamil, Telugu, Malayalam, and Kannada.

The three South Dravidian languages were a part of larger repertoire of the language identifiers presented by, for example, \citet{majlis1} and \citet{kocmi1}. \citet{hanumathappa1} conducted identification experiments between Kannada and Telugu, a Central Dravidian language.

\section{Shared Task Setup and Data}
\label{sec:Data}

The evaluation measure in the DLI shared task was the macro \(F_1\) score, which gives equal value for each language independent of their actual distribution in the test set.

The data set provided by the DLI organizers contains a total of 22,164 YouTube comments written in a mix of English and one of the aforementioned South Dravidian languages \citep{chakravarthi-etal-2020-sentiment,chakravarthi-etal-2020-corpus,hande-etal-2020-kancmd}. In addition to the target languages, the data included comments in other languages as well. It was divided into 16,674 instances for training and 4,590 instances for testing. The number of instances for each language is show in Figure \ref{tab:number_instances}.  

\begin{table}[!ht]
\centering
\setlength{\tabcolsep}{4.5pt}
\scalebox{0.92}{
\begin{tabular}{lccccc}
\hline
\bf Set  & \bf kan & \bf mal & \bf tam & \bf other & \bf Total \\ \hline

Training & 493 & 4,204 & 10,969 & 1,008 & 16,674   \\
Test &  &  &   & & 4,590 \\ \hline
\bf Total & & & & & 22,164 \\ \hline
\end{tabular}
}
\caption{Number of instances in the DLI dataset for Kannada (kan), Malayalam (mal), and Tamil (tam).}
\label{tab:number_instances}
\end{table}

\noindent In order to evaluate and compare our methods using the training data, we divided the training data into training and development portions. For training, we used the first 90\% of comments for each language and the rest was set aside for development. This way, we retained the original distribution of different labels as the provided training data seemed not to be in a random order. For example, the last 10\% of the training comments did not include comments in Malayalam at all.

For the open track, we considered using corpora collected for the use of the SUKI-project \cite{jauhiainen5}, but quickly found that for example the SUKI data for Tamil consisted of texts written completely using the Devanagari script. As the YouTube comments used in the shared task were mostly written in Latin alphabet, we did not pursue using external corpora further.

\section{Methods and Experiments with the development data}
\label{sec:Methods}

In our development environment, we experimented with several different methods: simple scoring, sum of relative frequencies, the product of relative frequencies (Naive Bayes, NB), NB with language model adaptation, HeLI, language set identification, and transformers.

\subsection{Simple scoring, sum of relative frequencies, and Naive Bayes}
\label{tunnistaexperiments}

None of the three methods are equipped with any special ways of handling multilingual texts and simply act in a similar fashion as each line would actually be monolingual. The possible multilinguality of the input texts is handled by the language models derived from the training material actually representing a mixture of languages.

The implementation of these methods was the same, which was used by \citet{jauhiainen2019mandarin} in the Discriminating between Mainland and Taiwan variation of Mandarin Chinese (DMT) shared task. We have now published this software titled ``Tunnista'' in GitHub with an MIT license.\footnote{\url{https://github.com/tosaja/Tunnista}} The software is used to automatically evaluate the performance of the methods using different ranges of character \emph{n}-grams and penalty modifiers.

In the evaluations of all three methods, we removed all non-alphabetical characters from the training and the test material and also lowercased all the remaining characters.

Simple scoring was evaluated with all possible character \emph{n}-gram ranges from 1 to 10 and from 6 to 11. The best micro-\(F_1\) score of 0.9225 was attained using character \emph{n}-grams from 7 to 8 and the best macro-\(F_1\) of \textbf{0.7321} with character \emph{n}-grams from 7 to 10. The sum of relative frequencies was evaluated with all possible character \emph{n}-gram ranges from 1 to 11. The best micro-\(F_1\) score of 0.8048 was attained using character \emph{n}-grams from 3 to 9 and the best macro-\(F_1\) of \textbf{0.6897} with character \emph{n}-grams from 7 to 9. It is noticeable how much closer the macro \(F_1\) scores are to each other when compared with the micro \(F_1\) scores.

The NB classifier was evaluated with several combinations of character \emph{n}-grams and penalty modifiers ranging from \emph{n}-gram length of 1 to 11 and penalty modifiers between 1.2 and 2.5. The classifier obtained its best micro \(F_1\) score of 0.9339 using character \emph{n}-grams from 1 or 2 to 6 with penalty modifiers ranging from 2.37 to 2.42. The best macro \(F_1\) score, \textbf{0.8609} was attained using character \emph{n}-grams from 2 to 6 with penalty modifiers ranging from 2.14 to 2.16. The NB classifier was clearly the best of the three methods provided by the ``Tunnista'' program.

\subsection{Product of Relative Frequencies with Adaptive Language Models}
\label{adaptiveexpperiments}

As the macro \(F_1\) score was used as the evaluation measure of the shared task, we used the best parameters (character \emph{n}-grams from 2 to 6, with penalty modifier of 2.15) from the previous experiments as the basis for our experiments with the adaptive version of the naive Bayes classifier.

The adaptation method uses several parameters which have to be optimized using the training and the development material. The first parameter is the number of splits the whole material to be identified is divided in. The actual division into splits happens after each time the test set is preliminarily identified and ordered so that the mystery texts with the highest difference between the log probabilities of the most probable and the second most probable language are on the top of the list. When incorporating new information from the text to be identified, the highest split is processed first. After its information has been added to the language models, all the remaining mystery texts are again preliminarily identified and divided into same sized splits. Again the information from the best split is incorporated into language models and so on, until all the splits have been processed.

We evaluated different split sizes between zero and the so called full split, which means the number of splits was equal to the number of mystery texts in the test set. Table~\ref{tab:koos} shows the development of the \(F_1\) score relative to the number of splits \(k\).

\begin{table}[h]
\center
\scalebox{0.92}{
\begin{tabular}{lc}
\hline
\bf \emph{k} & \bf macro \(F_1\)-score\\
\hline
1 & 0.8609\\
2 & 0.8595\\
4 & 0.8623\\
10 & 0.8648\\
\bf 20, 40, 100, 200, 400, 800, max & \bf 0.8663\\
\hline
\end{tabular}
}
\caption{The macro F1-scores obtained by the NB identifier using adaptation with different values of \(k\) when evaluated on the development partition of the DLI training set.}
\label{tab:koos}
\end{table}

\noindent We also use the log probability difference between the best and second best scores as a confidence score. In case confidence threshold, \(CT\), is used, no information from mystery lines with confidence score equal or below \(CT\) is incorporated into the language models. We evaluated several \(CT\) values with \(k\) of 20, but the best results were obtained without using the confidence threshold at all.

\subsection{HeLI 2.0 method}
\label{HelI2}

The third series of experiments was conducted using a language identifier based on the HeLI method \citep{jauhiainen-linden-jauhiainen:2016:VarDial3}. The actual implementation used was that of the HeLI 2.0 method \citep{jauhiainen2019nle} with adaptive language models from the GDI 2019 shared task \citep{jauhiainen2019mandarin}.\footnote{The publication of the HeLI 2.0 implementation is still currently in our queue.} We used the \(createmodels.java\)\footnote{\url{https://github.com/tosaja/HeLI}} program to generate language models of words and character \textit{n}-grams of up to 12 characters in length, both lowercased and with the original casing. All non-alphabetical characters were also removed.

Table~\ref{tab:heliexp} lists some combinations of parameters and the results they achieved for the development data in our experiments. Column heading \(lnr\) refers to the length range of lowercased character \textit{n}-grams, \(onr\) to those with original casing, \(lw\) and \(ow\) tell whether lowercased or origical words were used (y) or not (n). Column heading \(pm\) refers to the used penalty modifier.

From these experiments, it was clear that the naive Bayes based language identifier fared better than the one implementing the HeLI method. We also experimented with HeLI 2.0 using language model adaptation scheme identical to what we described in Section~\ref{adaptiveexpperiments}. These experiments failed to increase the macro \(F_1\) score at all.

\begin{table}[!ht]
\center
\scalebox{0.92}{
\begin{tabular}{lllllc}
\hline
\bf \emph{lnr} & \bf \emph{onr} & \bf \emph{lw} & \bf \emph{ow} & \bf \emph{pm} & \bf macro F1-score\\
\hline
1-11 & 1-11 & y & y & 1.17 & 0.8334\\
1-11 & 1-11 & y & n & 1.14 & 0.8305\\
1-11 & - & y & y & 1.18 & 0.8308\\
- & 1-11 & y & y & 1.17 & 0.8334\\
1-10 & 1-10 & y & y & 1.17 & 0.8334\\
1-9 & 1-9 & y & y & 1.17 & 0.8334\\
1-8 & 1-8 & y & y & 1.17 & 0.8334\\
1-7 & 1-7 & y & y & 1.10 & 0.8338\\
1-6 & 1-6 & y & y & 1.11 & \bf 0.8403\\
1-5 & 1-5 & y & y & 1.09 & 0.8369\\
2-6 & 2-6 & y & y & 1.11 & \bf 0.8403\\
3-6 & 3-6 & y & y & 1.10 & 0.8396\\
2-6 & 2-6 & y & n & 1.10 & 0.8319\\
2-6 & 2-6 & n & y & 1.11 & \bf 0.8403\\
\hline
\end{tabular}
}
\caption{The macro F1-scores obtained by HeLI-based identifier when evaluated on the development partition of the DLI training set. See Section~\ref{HelI2} for the explanations of the column headings. The best results are in bold.}
\label{tab:heliexp}
\end{table}

\subsection{Language Set Identification}

As the mystery lines in the DLI data set were actually multilingual, we wanted to experiment with a system capable of detecting several languages in one text. We set out to incorporate the language set identification method deviced by \citet{jauhiainen3} into the naive Bayes identifier with adaptive language models. A server version of the language set identifier using the HeLI method is currently available at GitHub.\footnote{\url{https://github.com/tosaja/TunnistinPalveluMulti}} It has been successfully used as part of corpus collection and creation pipeline resulting to the Wanca corpora, which has been used in the Uralic Language Identification (ULI) shared task \citep{jauhiainen2020building,ULIVarDial}. We managed to combine the method to the naive Bayes classifier and run some initial experiments, in which using the language set identification did not improve the results. Unfortunately, we did not have time to finalize our experiments as we were left contemplating about the nature of the multilinguality inherent in the DLI data set and were not encouraged by our initial results.

\subsection{Transformers}

The system for our second submission (described in Section~\ref{system2}) was based on pretrained transformer models: multilingual BERT \cite{devlin2018bert} and XLM-RoBERTa (XLM-R) \cite{conneau2019unsupervised}. The system used pretrained language models available from the Hugging Face Team \footnote{\url{https://huggingface.co/transformers/pretrained_models.html}}. Transformer are effective than RNN based deep learning architectures \cite{hettiarachchi-ranasinghe-2019-emoji} in text classification tasks \cite{ranasinghe2019brums, ranasinghe-hettiarachchi-2020-brums, pitenis2020}.

It was also evaluated using the same development set which was used in the previously described experiments. It the initial experiments it achieved a macro \(F_1\) score of 0.785, which was considerably lower than the 0.861 gained by the simple naive Bayes model even though it used pretrained models as opposed to the naive Bayes which was using only the data provided for the DLI task. We considered generating additional training material from the SUKI data and other available Dravidian corpora written with the native script using automatic transliteration.\footnote{Indic transliteration tools provide such functionalities - \url{https://pypi.org/project/indic-transliteration/}} However, we were doubtful about the performance of such libraries since the texts they were trained with were not natural romanized writing as opposed to the YouTube comments part of the DLI shared task and did not pursue this further.

\section{Submissions and Results}

In this Section, we summarize the systems we used in our submission and provide the results obtained on the test set.

\subsection{System 1: Bayesian Classifier}

The results of our first submission were produced using the adaptive version of the naive Bayes classifier with the best parameters attained in the experiments detailed in Sections~\ref{tunnistaexperiments} and \ref{adaptiveexpperiments}.

The system is based on a naive Bayes classifier using the relative frequencies of character \textit{n}-grams as probabilities \citep{jauhiainen2019mandarin}. Non-alphabetic characters were removed and the rest were lowercased. The lengths of the character \textit{n}-grams used were from two to six. Instead of multiplying the relative frequencies we summed up their negative logarithms. As a smoothing value we used the negative logarithm of an \textit{n}-gram appearing only once multiplied by a penalty modifier. In this case, the penalty modifier was 2.15.

In addition, we used identical language model adaptation technique as was used with the HeLI method in GDI 2018 \citep{jauhiainen9}. We used one epoch of language model adaptation to the test data. The \textit{n}-gram models used, the penalty modifier, the confidence value, and the number of splits in adaptation were optimized using 10\% of the training data as the development data.

\subsection{System 2: Transformers}
\label{system2}

These results were produced by using a pretrained transformer model which has been used in a number of NLP tasks like text classification \cite{pitenis2020}, span classification \cite{ranasinghemudes}, word similarity \cite{hettiarachchi-ranasinghe-2020-brums}, question answering \cite{yang-etal-2019-end} etc. We pass the sentence through the transformer model and add a softmax layer on top of the [\textsc{CLS}] token as a normal classification architecture with transformers \cite{ranasinghe-hettiarachchi-2020-brums}. We fine-tune all the parameters from transformer model as well as the softmax layer jointly by maximising the log-probability of the correct label. This architecture has been used widely in many text classification tasks \cite{ranasinghe2019brums, ranasinghe-etal-2020-multilingual} that includes Malayalam code-mix texts too \cite{ranasinghe2020}. We did not perform any preprocessing to this architecture. Considering the pretrained transformer models that supports Kannada, Malayalam and Tamil we used multilingual bert \cite{devlin2018bert} and XLM-R large \cite{conneau2019unsupervised} models. 

We divided the dataset into a training set and a validation set using 0.8:0.2 split on the data set. We predominantly fine tuned the learning rate and number of epochs of the classification model manually to obtain the best results for the validation set. We obtained $1e^-5$ as the best value for learning rate and 3 as the best value for number of epochs. We first fine-tune multiple transformer models with different random seeds. For each input, we output the best predictions made by the transformer along with the probability and sum up the probability of the predictions from each model together. The output of the ensemble model is the prediction with the highest probability. This voted ensemble method has improved results in many tasks \cite{hettiarachchi-ranasinghe-2020-infominer} for transformers. 

\subsection{Results on the Test Set}
\label{sec:results}

Table~\ref{DLIresults} shows the results of the shared task. Our first run was clearly better than the second and not far behind the results of the LAST-team.

\begin{table}[h]
\centering
\scalebox{0.92}{
\begin{tabular}{cccc}
\hline
\textbf{Rank} & \textbf{Team} & \textbf{Run} & \textbf{Macro \(F_{1}\)} \\
\hline
1 & LAST & 1 & 0.93 \\
 & LAST & 2 & 0.92\\
  & LAST & 3 & 0.92\\
\textbf{2} & \textbf{HWR} & \textbf{1} & \textbf{0.92}\\
2 & NYAEL & 1 & 0.92 \\
 & NAYEL & 2 & 0.91 \\
4 & Phlyers & 1 & 0.89 \\
 & Phlyers & 2 & 0.89 \\
 & \textbf{HWR} & \textbf{2} & \textbf{0.89} \\
 & NAYEL & 3 & 0.84 \\
\hline
\end{tabular}
}
\caption{The results of each team participating on the DLI shared task in terms of Macro F1. Our results are displayed in bold.}
\label{DLIresults}
\end{table}

\section{Discussion}
\label{sec:conclusion}

We did not experiment with the NB identifier using non-alphabetic characters nor with the original casing of the alphabetic characters. In light of the results for the HeLI method in Table~\ref{tab:heliexp}, it might be a promising direction. It seems that at least for the HeLI method, the original character \textit{n}-grams were more important than lowercased as can be seen from the third and fourth rows of the table. The same NB method was used also in the winning submission of the Romanian Dialect Identification (RDI 2021) shared task organized as part of the same campaign as DLI \citep{vardial2021report}. In RDI, the use of non-alphabetic characters and original casing proved out to be quite important \citep{jauhiainen2021} and might have very well cost us the small difference between the second and the first positions in the DLI shared task.

Even though the difference in performance between the NB model and the transformers was only 3 percentage points in the test set, the fact that the transformers did not outperform the simple naive Bayes classifier deserves special attention. One of the reasons to the inferior performance of the pretrained models is probably that the comments contained code-mixed sentences kind of which the pretrained language models like BERT and XLM-R had not seen before.

\section{Conclusion}

We present the submissions by team HWR to the Dravidian Language Identification (DLI) shared task at VarDial 2021. The DLI shared task featured a challenging dataset including YouTube comments written in Roman script containing code-mixed text with English and one of the three South Dravidian languages: Kannada, Malayalam, and Tamil. Our two systems, obtained competitive performance with the NB system achieving 2\textsuperscript{nd} position in the competition.

Our results are in line with the general trend of deep learning methods not being overtly competitive in language identification tasks as discussed in  \citet{medvedeva-kroon-plank:2017:VarDial}.

\section*{Acknowledgments}

We would like to thank the DLI organizers for making this interesting dataset available. 

This research has been partly funded by The Finnish Research Impact Foundation in cooperation with Lingsoft.

\bibliography{eacl2021}
\bibliographystyle{acl_natbib}

\end{document}